\documentclass[lettersize,journal]{IEEEtran}
\usepackage{amsmath,amsfonts}
\usepackage{algorithm}
\usepackage{array}
\usepackage[caption=false,font=normalsize,labelfont=sf,textfont=sf]{subfig}
\usepackage{textcomp}
\usepackage{stfloats}
\usepackage{url}
\usepackage{verbatim}
\usepackage{graphicx}
\usepackage{cite}
\usepackage[table,xcdraw]{xcolor}

\DeclareSymbolFont{matha}{OML}{txmi}{m}{it}
\DeclareMathSymbol{\varv}{\mathord}{matha}{118}

\usepackage{algpseudocode}
\usepackage{multirow}
\usepackage{cite}
\usepackage{amsmath,amssymb,amsfonts}
\usepackage{graphicx}
\usepackage{textcomp}
\usepackage{xcolor}
\usepackage{color}
\usepackage{algorithm}
\usepackage{algpseudocode}
\usepackage{authblk}
\usepackage{hyperref}
\usepackage{siunitx}

\usepackage{bm}
\usepackage{soul}

\IEEEoverridecommandlockouts                              

\providecommand{\keywords}[1]
{
  \small	
  \textbf{\textit{Keywords---}} #1
}

\title{\LARGE \bf
Affordance-Based Hierarchical Reinforcement Learning for Quadruped Pedipulation
}

\author{Tuba Girgin$^1$, Jose Castelblanco$^1$, Gabriel Rodriguez$^2$,  Emre Girgin$^1$, Cagri Kilic$^1$
\thanks{This study is partially supported by 2024-2025 COE-The Boeing Center for Aviation and Aerospace Safety (BCAAS)
Research Stimulus Program. }
\thanks{$^1$Department of Aerospace Engineering, Embry-Riddle Aeronautical University, Daytona Beach, FL, USA. \\
        { Email: \tt\small cibukgig@my.erau.edu}}%
\thanks{$^2$Robotics Institute, Carnegie Mellon University, Pittsburgh, PA, USA.}%
}

\begin{document}

\maketitle
\thispagestyle{empty}
\pagestyle{empty}

\begin{abstract}

The object manipulation capabilities of quadruped robots is an open research challenge. While previous studies have focused on low-level policy learning, task execution still relies on expert-designed high-level trajectories. Autonomous selection of both an affordable interaction point on the target object and an affordable robot base pose removes the need for pre-designed trajectories.  This study proposes a three-level hierarchical reinforcement learning (RL) framework that utilizes pose affordances to guide the navigation policy, while the navigation policy drives the locomotion policy. In addition, the pedipulation policy is guided by interaction-point affordances, enabling object-centric pose alignment of the quadruped robot and effective end-effector manipulation planning. We train the proposed framework in the IsaacSim ecosystem and evaluate it in both simulation and real-world settings. We investigate the effectiveness of pose affordance across multiple scenarios in simulation while various object interaction tasks are validated on real-world setting forming an object-interaction dataset. 
The results show that the proposed framework can autonomously identify candidate poses based on their affordance and successfully execute object manipulation tasks in the real world without human guidance.

\end{abstract}

\keywords{Affordance-based planning, hierarchical reinforcement learning, quadruped robots, pedipulation, autonomous manipulation}

\section{INTRODUCTION}


Autonomous robots are highly utilized in inspection, rescue and recovery operations, and planetary exploration missions, where human guidance or intervention is dangerous or not available, particularly in scenarios where real-time teleoperation is infeasible due to communication latency \cite{rankin2021mars}. In such applications, the autonomy of quadruped robots has been extensively studied due to their superior locomotion capabilities \cite{qi2024reinforcement,ji2021reinforcement}. Reinforcement learning has been widely employed to extend the skill set of quadruped robots, focusing not only on advanced locomotion \cite{wang2024learning} but also on manipulation tasks through whole-body \cite{jeon2023learning}, robotic arms \cite{fu2023deep, jiang2024learning,liu2025mlm}, or limb-based \cite{stolle2024perceptive},  where manipulation using the legs is referred to as pedipulation. Recent studies on object manipulation \cite{10161470,arm2024pedipulate} focus on low-level policy learning, but experiments rely heavily on expert guidance. In contrast, end-to-end object-centric pedipulation planning in open-world environments, including the identification of afforded base poses to start pedipulation, is underexplored. We address this gap in the literature.

Manipulating objects on uneven terrain introduces additional challenges \cite{hu2026towards}, such as unmodelled surface geometry and unknown surface friction. Although limb-based object manipulation is the main objective, reaching a feasible  pose and selecting contact points are crucial for successful operation. Furthermore, maintaining balance on the remaining limbs during object interaction becomes particularly challenging when the properties of both the manipulated object and the terrain are unknown.

\begin{figure*}[t]
    \centering
    \includegraphics[ width=1.0\linewidth, trim={0 0 0 0}, clip]{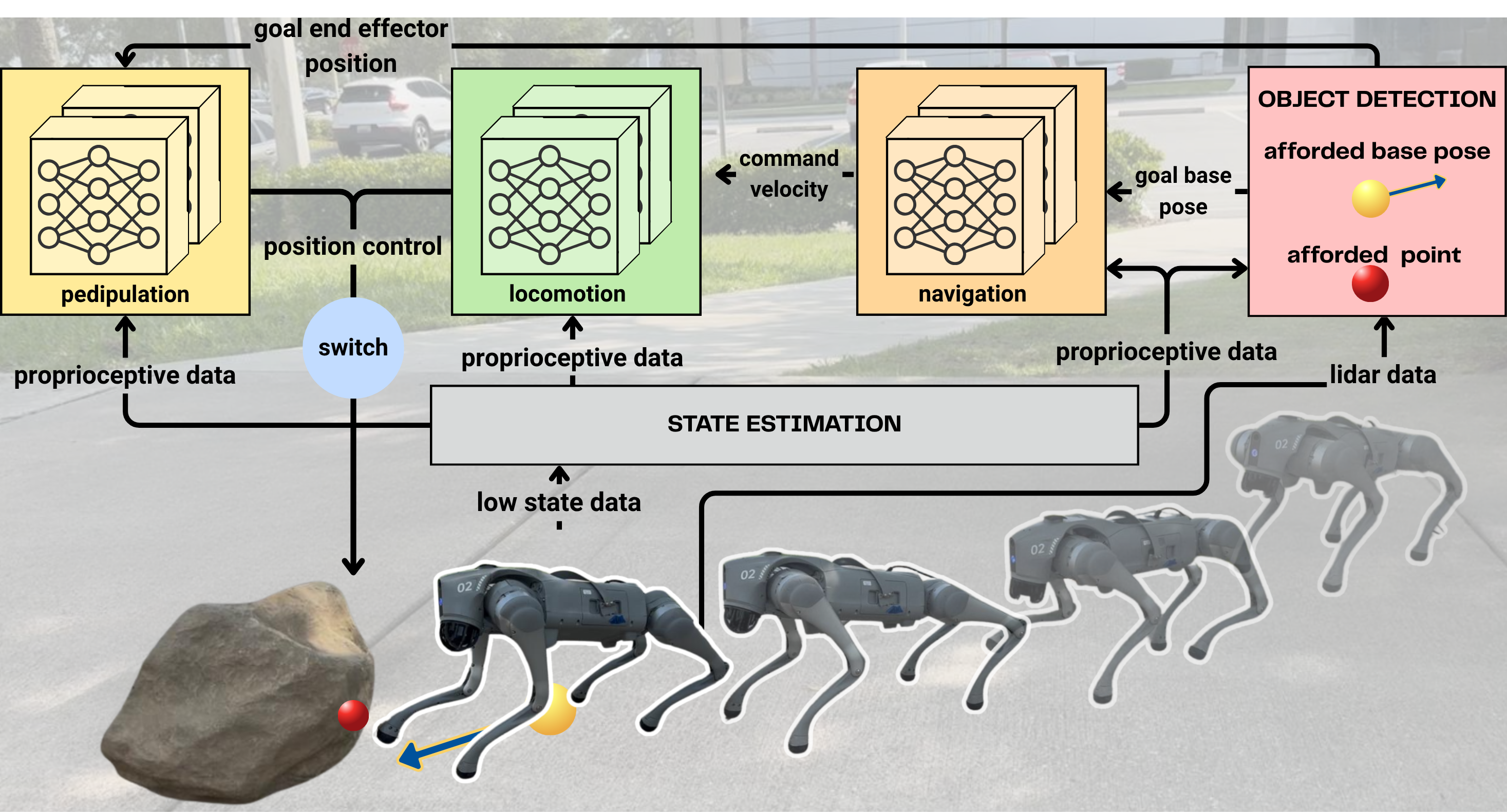}
    \caption{ An illustration of the three-level hierarchical reinforcement learning framework. Our vision-based robot pose affordance model processes the point cloud data acquired from the LiDAR of the Go2 quadruped robot, along with odometry from the state estimation module. It generates a goal base pose for the navigation policy's observation space. The navigation policy receives linear velocity and robot orientation from the state estimation module and predicts the command velocity. The locomotion policy then outputs joint positions for the robot, conditioned on the command velocity and the provided robot state. Once the goal pose is reached within a defined threshold, the system switches to pedipulation, conditioned on the identified interaction point.}
    \label{fig:pipeline}
\end{figure*}

In this study, we propose a three-level hierarchical reinforcement learning (RL) framework visualized in Figure \ref{fig:pipeline}. At the highest level, a vision-based pose affordance module processes the point cloud data acquired from the quadruped robot’s LiDAR sensor and generates an affordable goal pose for the robot. Following terrain and object segmentation on point cloud, the local terrain slope is calculated to determine an affordable robot heading as in previous work \cite{girgin2025learning}. The goal position is subsequently computed based on the reachability constraints of the end effector. The goal pose is then fed to the navigation policy, which predicts velocity commands to guide the low-level locomotion policy in a closed-loop manner. The decoupled architecture of the proposed framework enables the integration of pretrained models in a plug-and-play manner. For the pedipulation task, the high-level vision-based pose affordance module further identifies an initial contact point between the end-effector and the object by analyzing the surface geometry of the object. The low-level pedipulation policy is subsequently conditioned on high-level waypoints composed of target end-effector poses generated using cubic spline interpolation between the home position and the afforded interaction points. Eventually, the proposed framework autonomously determines feasible robot base poses for end-effector interaction with the target object, identifies candidate interaction points through vision-based affordance analysis, navigates toward the target pose driven by navigation and locomotion policies, and executes object interaction through the pedipulation policy, all without requiring human demonstrations or manual guidance.

\begin{figure}
    \centering
    \includegraphics[ width=0.5\textwidth, trim={0 0 0 0}, clip]{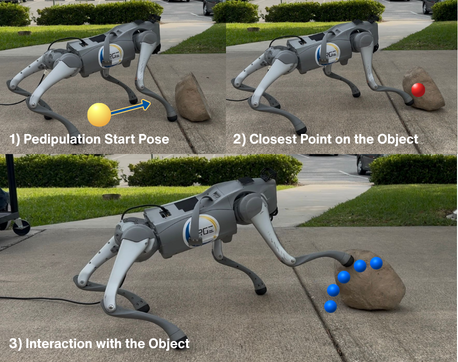}
    \caption{Pedipulation of the target object. After reaching the goal base pose (yellow sphere), the end effector (the robot’s front right foot) interacts with the object at the identified feasible interaction point (red sphere). The high-level end-effector waypoints are represented by the blue spheres in the third image.}
    \label{fig:robot_sim}
\end{figure}

We train the RL policies in the Isaac Sim ecosystem using the Unitree Go2 quadruped robot platform. The proposed framework was evaluated in both simulation and outdoor real-world environments, where the robot was tasked with interacting with randomly positioned target objects on varying surface slopes. The task design was inspired by biological systems, particularly the way cats interact with surrounding objects through pedipulation to explore and infer object properties \cite{hall1998object}. This behavior closely aligns with planetary exploration and sample collection scenarios as \cite{arm2023scientific} uses robotic arms on quadrupeds to inspect surrounding terrain and rocks. During navigation towards the goal base pose and throughout the pedipulation routine, we collected multimodal data, including object features, target base poses, target pedipulation poses, robot base odometry, joint positions, and force feedback generated during object interaction. These data were used to construct a dataset for open-world object interaction using the limbs of quadruped robots.





The main contributions of this study are as follows:
\begin{enumerate}
    \item A visual-affordance-based pose command generation module for both the quadruped robot base and the end-effector, enabling affordable object interaction while accounting for object reachability, local terrain characteristics, and object surface features.
    
    \item A multi-level hierarchical RL framework that enables fully autonomous quadruped navigation and pedipulation using the end effector, while abstracting the behavior of the other layers as black boxes. 
    
    \item A sim-to-real implementation demonstrating the transferability of the proposed system in open and unstructured real-world environments.
    
    \item An object interaction dataset containing the robot telemetry, including force feedback from the end effector, along with extracted object features, thereby enabling further investigation of object exploration and interaction tasks.
\end{enumerate}

The remainder of this paper is organized as follows. Section~\ref{sec:related_work} reviews the related work. Section~\ref{sec:methodology} presents the proposed methodology. Section~\ref{sec:experiments} describes both the simulation and real-world experiments with their results. Finally, Section~\ref{sec:conclusion} concludes the paper and discusses the insights gained during the experimental studies.

\section{RELATED WORK}
\label{sec:related_work}

With recent advancements in reinforcement learning for manipulation with quadruped robots, whole-body interaction and locomotion-based manipulation have been extensively studied in the literature \cite{shi2021circus, sombolestan2024hierarchical, zhang2025manipulate}. Azimi et al. \cite{azimi2025hierarchical} focus on whole-body object manipulation using a hierarchical reinforcement learning framework with sensor-based reward design. In their approach, the high-level navigation policy receives estimated object and target positions and outputs desired base velocities for the low-level locomotion policy. The reward function also considers safety constraints by penalizing behaviors that bring the robot too close to obstacles such as walls or result in unsafe heading angles. Similarly, Jeon et al. \cite{jeon2023learning} propose a hierarchical reinforcement learning framework for whole-body manipulation of heavy objects. Their method leverages privileged information available during training in simulation, which is encoded into a latent representation by the high-level policy. This latent space enables the policy to implicitly capture object properties and improve manipulation performance under partial observability. Feng et al. \cite{feng2025learning} introduce a multi-agent hierarchical reinforcement learning framework composed of three levels of control. A global rapidly exploring random tree (RRT) based planner first generates high-level trajectories. The high-level controller then produces subgoals conditioned on the state of the environment and interacting robots. Based on these subgoals, mid-level policies generate velocity commands for low-level controllers. The mid-level policies additionally incorporate observations of other agents to enable coordinated behavior.

While studies on whole-body manipulation focus on using the robot’s body for object interaction, other works explore dexterous manipulation strategies that rely on dedicated end-effectors for precise interaction with objects. Farid et al. \cite{farid2023simultaneous} propose a radial basis function neural network-based actor–critic policy for quadruped manipulation, enabling the control of joint trajectories and interaction forces. Their approach is grounded in Lyapunov stability theory, providing a proof of uniform ultimate boundedness (UUB) of the tracking errors. However, the evaluation is limited to simulation studies.

Equipping a quadruped robot, which already possesses four 3-Degree of Freedom (DOF) legs, with an additional end effector introduces increased system complexity, higher payload demands, and redundancy in the kinematic structure. Cheng et al. \cite{10161470} utilize the robot’s legs as manipulators, learning a low-level manipulation policy together with a locomotion policy. On top of these skills, they learn a behavior tree that enables switching between policies based on human demonstrations. Arm et al. \cite{arm2024pedipulate} learn a low-level pedipulation policy for the front right leg to achieve commanded end-effector target positions, while acquiring tripod locomotion capabilities. Yet, high-level task execution is performed via a manual joystick interface. Stolle et al. \cite{stolle2024perceptive} extend this framework by incorporating obstacle avoidance through a contact-switch mechanism for the end effector, while still primarily focusing on low-level control policy improvements. He et al. \cite{he2024learning} propose a behavior cloning-based high-level planner built on top of a reinforcement learning-based low-level controller. Their approach leverages expert demonstrations and conditions the policy on visual point cloud inputs to generalize across different task scenarios. Zhang et al. \cite{zhang2025bipedalism} introduce risk-adaptive distributional reinforcement learning for bipedal locomotion, enabling manipulation capabilities using the front limbs during locomotion. While these approaches focus on locomotion-centric or low-level control strategies, our work emphasizes \emph{object-centric pedipulation}, explicitly considering the state and geometry of the object to compute feasible and affordance-aware interaction poses for successful manipulation.

The concept of affordance, originally introduced by Gibson \cite{gibson1977theory}, has been widely adopted in robotics and is increasingly studied in quadruped robotic systems for feasible trajectory and interaction point planning \cite{escontrela2025learning}. Saputra et al. \cite{saputra2021aquro, saputra2023neuro} propose a bio-inspired attention-based topological mapping approach to identify affordance-rich regions in the environment, enabling the detection of feasible stepping points for locomotion as well as safe graspable points for ladder climbing tasks.



While reinforcement learning-based studies primarily focus on low-level control policies and expert-guided trajectory learning, and affordance-based approaches are mainly centered on improving locomotion through the identification of feasible stepping points, there remains a gap in affordance-driven manipulation. In this work, we address this gap by proposing an affordance-based goal pose generation framework that enables reachability-aware interaction with target objects using the legs of a quadruped robot.

\section{METHODOLOGY}
\label{sec:methodology}


\subsection{Problem Formulation for Reinforcement Learning}

We define the navigation, locomotion, and pedipulation tasks each as a Markov Decision Process (MDP) represented by a 5-tuple  $\langle\mathcal{S, A, P, R,} \gamma\rangle$. The state space $S$ generally consists of proprioceptive sensor measurements and control commands, the latter varying on the policy's level in the architecture. The terms state and observation are used interchangeably in the following parts. The action space $\mathcal{A}$ includes position-controlled joint commands for the low-level policies, while for the high-level navigation policy, the action space corresponds to base position commands. The transition probability $\mathcal{P}$ is governed by the dynamics of the Isaac Sim simulation environment. Similarly to $\mathcal{S}$, the reward function $\mathcal{R}$ is defined differently for each task to reflect their objectives, and is described in the following subsections. The discount factor, $\gamma$, used to calculate the total accumulated reward.



\subsubsection{Pedipulation Policy}
The state space of the pedipulation policy consists of the robot base frame's 3D linear velocity, 3D angular velocity, 1D base height in the world frame, the 3D gravity vector expressed in the base frame, a 7D end-effector pose command (3D position in the base frame and 4D quaternion orientation), 12D joint positions, 12D joint velocities, and the 12D previous action. In total, the observation space of the policy is represented by a 53-dimensional vector, the result of flattening the state vectors. The action space consists of a 12D vector of target joint positions, $\bm{\theta} = [\theta_{1}, \dots, \theta_{12}]^T,$ where the robot is controlled using joint position control in simulation.


The reward function for the pedipulation policy, denoted by $R^p$, is defined as the sum of the end-effector position tracking reward $R_g^p$ and the penalty terms $R_p^p$, where




\begin{equation}
\label{eq:reward_function}
\begin{aligned}
R^p &= R_g^p + R_p^p, \\
R_g^p &= c_b R_b^p + c_{ee} R_{ee}^p, \\
R_b^p &=
\begin{cases}
1, & \text{if } \lVert \bm{p}_{ee} - \bm{p}_{ee,g} \rVert < \delta_{ee} \;\land\; t_r \geq t_{\min}, \\
0, & \text{otherwise},
\end{cases} \\
R_{ee}^p &=
\frac{1}{1 + \lVert \bm{p}_{ee} - \bm{p}_{ee,g} \rVert}
+
\left(
1 -
\tanh\left(
\frac{\lVert \bm{p}_{ee} - \bm{p}_{ee,g} \rVert}{\sigma_{ee}}
\right)
\right), \\[5pt]
R_p^p &= 
c^p_v \lVert ^\mathcal{B}\!\bm{v} \rVert^2
+
c^p_\omega \lVert ^\mathcal{B}\!\bm{\omega} \rVert^2
+
c^p_\tau \lVert \bm{\tau} \rVert^2 \\
&\quad +
c^p_a \lVert \bm{a} \rVert^2
+
c^p_r \lVert \bm{\theta}_t - \bm{\theta}_{t-1} \rVert^2 .
\end{aligned}
\end{equation}

The goal reward $R_g^p$ is defined as the weighted sum of the bonus reward $R_b^p$ and the end-effector position tracking reward $R_{ee}^p$. The bonus reward $R_b^p$ is activated when the end-effector position $\bm{p}_{ee} \in \mathbb{R}^3$ remains within a threshold distance $\delta_{ee}$ of the goal position $\bm{p}_{ee,g} \in \mathbb{R}^3$ for a minimum duration $t_{\min}$, where $t_r$ denotes the accumulated time spent within the threshold region. The tracking reward $R_{ee}^p$ encourages minimization of the end-effector position error, while the scaling factor $\sigma_{ee}$ controls the smoothness of the reward shaping. The penalty term $R_p^p$ discourages excessive base linear and angular velocities $\left(^\mathcal{B}\!\bm{v}, ^\mathcal{B}\!\!\bm{\omega}\in \mathbb{R}^3\right)$, joint torques, joint accelerations, and action rate variations $\left( \bm{\tau}, \bm{a}, \bm{\theta} \in \mathbb{R}^{12}\right)$. The coefficients $c_*$ denote the corresponding weighting factors for each reward component.

\subsubsection{Locomotion Policy} 

The state space of the locomotion policy differs from the pedipulation policy in the command vector, which consists of the desired base velocity command, comprised of goal linear velocities along the $x$- and $y$-axes $^\mathcal{B}\bm{v}^{x,y}_g$, and a goal angular velocity about the $z$-axis $^\mathcal{B}\!\omega^z_g$. Consequently, the observation space is represented by a 49-dimensional vector. The action space remains identical to the pedipulation policy and consists of 12D target joint positions $\bm{\theta}$. The reward function for the locomotion policy, denoted by $R^l$, is composed of the velocity tracking reward $R_t^l$, the feet air-time reward $R_f^l$, and the penalty term $R_p^l$, where

\begin{equation}
\label{eq:locomotion_reward}
\begin{aligned}
R^l &= R_t^l + R_f^l + R_p^l, \\[4pt]
R_t^l &=
c_t
\exp\left(
-\frac{
\lVert ^\mathcal{B}\bm{v}^{x,y} - ^\mathcal{B}\bm{v}_{g}^{x,y} \rVert^2
}{
{\sigma_l}^2
}
\right)
+
c_o
\exp\left(
-\frac{
|^\mathcal{B}\!\omega^z - ^\mathcal{B}\!\omega_g^z|^2
}{
{\sigma_l}^2
}
\right), \\[6pt]
R_f^l &=
\begin{cases}
c_f
\displaystyle\sum_{i=1}^{4}
\left(
(s_i - t_h^l) f_{c_i}
\right),
& \text{if } \lVert ^\mathcal{B}\bm{v}_g^{x,y} \rVert \geq 0.1, \\[10pt]
0,
& \text{otherwise},
\end{cases} \\[12pt]
R_p^l &=
c_v^l (^\mathcal{B}\!v^z)^2
+
c_{\omega}^l \lVert ^\mathcal{B}\!\bm{\omega}^{x,y} \rVert^2
+
c_{\tau}^l \lVert \bm{\tau} \rVert^2 \\
&\quad +
c_a^l \lVert \bm{a} \rVert^2
+
c_r^l \lVert \bm{\theta}_t - \bm{\theta}_{t-1} \rVert^2
+ c_g^l
\lVert ^\mathcal{B}\!\bm{g}^{x,y} \rVert^2 .
\end{aligned}
\end{equation}


The tracking reward $R_t^l$ encourages tracking of the commanded linear velocity $^\mathcal{B}\bm{v}^{x,y}_{g}$ and yaw angular velocity $^\mathcal{B}\!\omega^z_{g}$. The scaling factor $\sigma^l$ determines the sensitivity of the tracking reward to velocity tracking errors. The feet air-time reward $R_f^l$ encourages longer swing phases by rewarding each foot upon its first ground contact, where $\bm{f}_c \in \mathbb{R}^4$ represents the foot first contact indicators, and $\bm{s} \in \mathbb{R}^4$ represents the feet air-time durations. This reward is activated only when the commanded velocity magnitude exceeds a standing threshold. The parameter $t_h^l$ defines the desired air-time duration threshold for each foot. Similar to the pedipulation policy, the penalty term $R_p^l$ discourages excessive body motion, joint torques, joint accelerations, action rate changes, and deviations in body orientation. Specifically, only the vertical component of the base linear velocity and the planar components of the base angular velocity are penalized. Additionally, the planar components of the projected gravity vector $^\mathcal{B}\!\bm{g} \in \mathbb{R}^3$  in the base frame are penalized to encourage stable body orientation.

The trained locomotion policy is used solely as a pretrained policy during navigation policy training and in simulation experiments. For real-world experiments, the trained navigation policy operates in conjunction with the robot's built-in locomotion controller.

\subsubsection{Navigation Policy} 

The state space of the navigation policy consists of the robot base-frame 3D linear velocity, the 3D gravity vector expressed in the base frame, and a 4D pose command in the base frame, resulting in a 10-dimensional observation vector. The action space is a 3D high-level command vector $(^\mathcal{B}\!v^x_g, ^\mathcal{B}\!\!v^y_g, ^\mathcal{B}\!\!\omega^z_g)$, which is provided as input to a pretrained low-level locomotion policy. The reward for the navigation policy, denoted by $R^n$, is composed of a position tracking reward $R_t^n$ and an orientation tracking reward $R_w^n$, where

\begin{equation}
\label{eq:navigation_reward}
\begin{aligned}
R^n &= R_t^n + R_w^n, \\[6pt]
R_t^n &= c_t \left(
\left(1 - \tanh\left(\frac{\lVert ^\mathcal{B}\bm{p}_g^{x,y} \rVert}{\sigma_{1}^n}\right)\right)
+
\left(1 - \tanh\left(\frac{\lVert ^\mathcal{B}\bm{p}_g^{x,y} \rVert}{\sigma_{2}^n}\right)\right)
\right), \\[10pt]
R_w^n &= c_o \left| ^\mathcal{B}\!\!\omega^z_g \right|.
\end{aligned}
\end{equation}

While $R_t^n$ rewards the reduction of the planar position ($ ^\mathcal{B}\bm{p}_g^{x,y} \in \mathbb{R}^2$) error in the base frame, thereby encouraging the robot to approach the desired target position, $R_w^n$ penalizes large commanded yaw values. This formulation implicitly encourages smaller orientation commands as the robot approaches the goal.


\begin{figure*}
    \centering
    \includegraphics[ width=1.0\linewidth, trim={0 0 0 0}, clip]{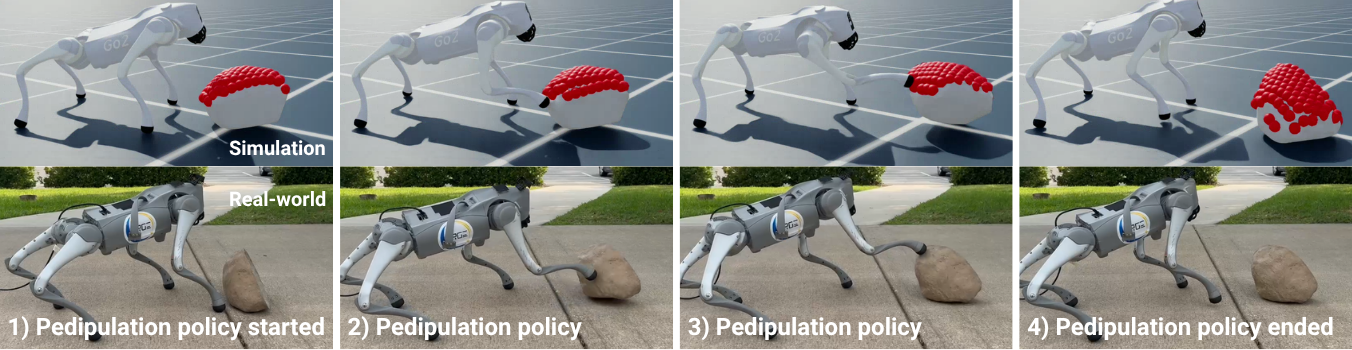}
    \caption{Still frames of the execution of the pedipulation operation in both simulation (first row) and real-world outdoor experiments (second row). Similar results verify the sim-to-real transfer and overall framework.}
   \label{fig:comp}
\end{figure*}

\subsection{Problem Formulation for Pose Affordance}

The objective is to generate afforded robot base poses $^\mathcal{B}\mathcal{P} = (^\mathcal{B}\bm{p}_g^{x,y}, \psi_g)$ as commands for the navigation policy to enable improved pedipulation of the target object while accounting for terrain constraints. After reaching the desired base position, the surface geometry of the target obstacle is exploited to compute an affordance-based contact point on the object surface, which serves as the pedipulation goal. In the real-world experiments, obstacles are detected and localized following the method proposed in \cite{girgin2025learning} (further details are provided in Section~\ref{real_world_section}). Algorithm~\ref{alg:robot_base_goal} describes the procedure for generating affordance-based goal base poses as commands to the navigation policy.

\begin{algorithm}[]
\caption{Robot Base Pose Goal}
\label{alg:robot_base_goal}
\begin{algorithmic}[1]
\Require $\psi$, $^\mathcal{W}\bm{p}_{robot}$, $^\mathcal{W}\!\bm{q}_{robot}$, $^\mathcal{W}\!\bm{o}$, $^\mathcal{B}\bm{p}_{ee,d}$, $^\mathcal{W}\!\bm{z}$, $\bm{q}_{ee}$

\Ensure $^\mathcal{B}\mathcal{P} = (^\mathcal{B}\bm{p}_g^{x,y}, \psi_g) $ 

\State $\bm{q}_{01}
\leftarrow
\operatorname{quat}(\psi, ^\mathcal{W}\!\bm{z})$

\State $(^\mathcal{W}\bm{p}_g, \bm{\_})
\gets
\operatorname{ComposeTransform}
(^\mathcal{W}\!\bm{o}, \bm{q}_{01}, ^\mathcal{B}\bm{p}_{ee,d}, \bm{q}_{ee})$

\State $^\mathcal{B}\bm{p}_g^{x,y}
\gets
\operatorname{QuatApplyInverse}
\left(
^\mathcal{W}\bm{q}_{robot},
^\mathcal{W}\bm{p}_g - ^\mathcal{W}\bm{p}_{robot}
\right)$

\State $\psi_{robot}
\gets
\operatorname{YawFromQuat}(^\mathcal{W}\bm{q}_{robot})$

\State $\Delta \psi \gets \psi - \psi_{robot}$
\State $\psi_g
\gets
(\Delta \psi + \pi) \bmod (2\pi) - \pi$

\State \Return $^\mathcal{B}\mathcal{P}$
\end{algorithmic}
\end{algorithm}

The position and quaternion of the robot base frame, expressed in the world frame, are denoted by $^\mathcal{W}\!\bm{p}_{robot}$ and $^\mathcal{W}\!\bm{q}_{robot}$, respectively. The desired pushing heading, represented by $\psi$, is adopted from \cite{girgin2026non}, and $^\mathcal{W}\!\bm{o}$ denotes the target object center position expressed in the world frame. The vector $^\mathcal{B}\bm{p}_{ee,d} \in \mathbb{R}^3 $ defines the desired end-effector position in the base frame, computed as the mean of the 3D domain defined during training. Furthermore, $\bm{q}_{ee}$ denotes the desired end-effector quaternion with respect to the body frame, and $^\mathcal{W}\!\bm{z} \in \mathbb{R}^3 $ is the unit vector indicating the positive direction of the Cartesian $z$-axis.

Given desired pushing heading $\psi$ and positive direction of the Cartesian $z$-axis $^\mathcal{W}\!\bm{z}$, goal quaternion $\bm{q}_{01}$ is calculated. Then, with the object center position $^\mathcal{W}\!\bm{o}$, calculated goal quaternion $\bm{q}_{01}$, and desired end-effector position $^\mathcal{B}\bm{p}_{ee,d}$ and orientation $\bm{q}_{ee}$, the desired goal position in the world frame $^\mathcal{W}\bm{p}_g$ is computed using standard composition. Using the current robot base pose $(^\mathcal{W}\bm{p}_{robot}, ^\mathcal{W}\bm{q}_{robot})$, this goal pose is then transformed into the robot base frame, discarding the z-axis component.

The yaw goal $\psi_g$ is computed as the difference between the current robot heading derived from $^\mathcal{W}\bm{q}_{robot}$ and the desired goal heading $\psi$, normalizing it in the range of $(-\pi,+\pi)$. The resulting goal pose is provided as input to the navigation policy. We assume that no dynamic obstacles are present, allowing the goal pose to be computed once to reduce noise induced by per-step sensor fluctuations. However, the proposed formulation is capable of handling dynamic goal updates if required.

\begin{algorithm}[]
\caption{Pedipulation Spline Creation and Sampling}
\label{alg:pedipulation_spline}
\begin{algorithmic}[1]
\Require $\bm{p}_{home}$, $\bm{p}_{push}$, $T$,$T_1$,$T_2$ ,$t_{now}$, $t_{start}$, $\delta_h$
\Ensure $\bm{p}_{cmd}$

\State $\bm{p}_{offset}
\gets
\bm{p}_{push}
+
\begin{bmatrix}
0.15 & -0.15 & 0
\end{bmatrix}^{T}$

\State $\boldsymbol{t}
\gets
\begin{bmatrix}
0 & T_1 & T_2 & T
\end{bmatrix}^{T}$

\State $\bm{W}
\gets
\begin{bmatrix}
\bm{p}_{home}^{T} \\
\bm{p}_{push}^{T} \\
\bm{p}_{offset}^{T} \\
\bm{p}_{home}^{T}
\end{bmatrix}$

\State $S(t)
\gets
\operatorname{CubicSpline}
\left(
\boldsymbol{t},
\bm{W},
\dot{S}(0)=\bm{0},
\dot{S}(T)=\bm{0}
\right)$

\State $t \gets t_{now} - t_{start}$

\State $\bar{t}
\gets
\operatorname{Clamp}(t, 0, T)$

\State $\bm{p}_{cmd}
\gets
S(\bar{t})$

\State $p_{cmd,z}
\gets
\min(p_{cmd,z},\delta_h)$

\State \Return $\bm{p}_{cmd}$
\end{algorithmic}
\end{algorithm}

Algorithm~\ref{alg:pedipulation_spline} describes the generation of position commands for the end-effector; $\bm{p}_{home} = (^\mathcal{B}\!p^x_{home}, ^\mathcal{B}\!\!p^y_{home}, ^\mathcal{W}\!\!p^z_{home})$ denotes a predefined safe home position to which the robot defaults when no active pedipulation is executed, $\bm{p}_{push} = (^\mathcal{B}\!p^x_{push}, ^\mathcal{B}\!\!p^y_{push}, ^\mathcal{W}\!\!p^z_{push})$ represents the target contact point on the object surface estimated by the vision module, as described in Section~\ref{real_world_section}. While the $x$- and $y$-components are expressed in the base frame, the $z$-component is expressed in the world frame to make the resulting end-effector height independent of the body height, which changes while pedipulation is being executed. The total duration of the pedipulation motion is denoted by $T$, $T_1$ represents one-third of the total duration, and $T_2$ represents two-thirds of the total duration, $t_{now}$ the current time, and $t_{start}$ the onset time of the pedipulation phase. An intermediate offset point $\bm{p}_{offset}$ is computed to guide the object along an arc-shaped trajectory toward the front-right direction of the leg. A cubic spline trajectory is then constructed using the defined waypoints, with zero velocity constraints imposed at both the initial and terminal points, ensuring a smooth transition from and to the home position. The spline is computed once at initialization and subsequently queried through time-based sampling. For safety, the maximum height of the resulting trajectory is clamped to a predefined limit $\delta_h$, which is 7.5 cm in our experiments, to prevent unexpected jumps in height because of the nonlinearity of fitted spline. The resulting end-effector reference trajectory is finally provided as the goal input to the pedipulation policy.

Eventually, once an object is detected for interaction, Algorithm~\ref{alg:robot_base_goal} generates navigation goals by incorporating the local environment around the object. The navigation policy then commands the low-level locomotion policy to reach the desired base pose. After the goal position is reached, Algorithm~\ref{alg:pedipulation_spline} is executed to generate end-effector trajectories for the pedipulation policy, enabling interaction with the target object based on its observable surface geometry.

\section{Experiments}
\label{sec:experiments}

Since existing pedipulation studies are not publicly available, we benchmark our approach under a variety of conditions to evaluate its effectiveness. We first compare the displacement achieved per physical interaction across different interaction directions and demonstrate, in simulation, the advantages of selecting an appropriate contact point on the obstacle surface. We then extend our evaluation to real-world experiments to assess both sim-to-real transfer performance and the effectiveness of the complete pipeline under non-ideal operating conditions.

\subsection{Simulation Experiments}


\subsubsection{Training}


\begin{table}
\centering
\caption{PPO Algorithm Hyperparameters}
\label{tab:ppo_hyperparameters}
\begin{tabular}{|l|c|}
\hline
\textbf{Hyperparameter} & \textbf{Value} \\ \hline
Learning Rate & $1.0 \times 10^{-3}$ \\
Learning Rate Schedule & Adaptive \\
Discount Factor ($\gamma$) & 0.99 \\
GAE Parameter ($\lambda$) & 0.95 \\
Clip Parameter ($\epsilon$) & 0.2 \\
Entropy Coefficient & 0.005 \\
Value Loss Coefficient & 1.0 \\
Desired KL Divergence & 0.01 \\
Number of Epochs & 5 \\
Number of Mini-batches & 4 \\
Max Gradient Norm & 1.0 \\ \hline
\end{tabular}
\end{table}

The policies are designed and trained using an actor–critic architecture with Proximal Policy Optimization (PPO) \cite{schulman2017proximal} within Isaac Lab \cite{mittal2025isaac}. For all trained models, both the actor and critic networks consist of three fully connected layers with a hidden layer size of 128. The exponential linear unit (ELU) is used as the activation function between layers. The PPO hyperparameters are summarized in Table~\ref{tab:ppo_hyperparameters}. Each model is trained for 4000 epochs to ensure convergence of torque and acceleration penalty terms.

For the pedipulation policy, target pose commands are uniformly sampled for the right end-effector within the range $x \in [0.35, 0.55]$ \SI{}{\meter}, $y \in [-0.30, -0.10]$ \SI{}{\meter}, and $z \in [0.00, 0.30]$ \SI{}{\meter}, while the orientation is fixed to the identity quaternion $[1, 0, 0, 0]$. For the left end-effector, the $y$-axis range is mirrored to positive values to maintain symmetry. For the locomotion policy, the base velocity command space is defined as $v_x, v_y \in [-1.0, 1.0]$ \SI{}{\meter\per\second} and yaw heading $\in [-\pi, \pi]$ \SI{}{\radian}. For the navigation policy, the pose command space is defined as $x, y \in [-3.0, 3.0]$ \SI{}{\meter} and heading $\in [-\pi, \pi]$ \SI{}{\radian}. 

During training, domain randomization is applied by sampling additional base mass variations $\Delta m \in [-5.0, 5.0]$ \SI{}{\kilogram}, and by randomizing the center of mass within $\pm 0.05$ \SI{}{\meter} along the $x$ and $y$ axes and $\pm 0.01$ \SI{}{\meter} along the $z$ axis. In addition, external perturbations are introduced during training in the form of random external forces, torques, and base velocity disturbances to improve robustness.

\subsubsection{Testing}

 In order to evaluate the proposed framework, we design an end-to-end task in which a GO2 quadruped robot is commanded to manipulate and push a target obstacle. In this setup, a goal pose is first generated using the pose affordance algorithm. The navigation policy then drives the locomotion policy by producing velocity commands to reach the desired goal pose. Once the robot reaches the target base position within a predefined threshold, the pedipulation policy is activated. The corresponding end-effector reference trajectory is generated using a cubic spline constructed from four key waypoints, (1) the home position of the end effector, (2) the vision-based contact point on the object surface, (3) an intermediate offset point, and (4) the home position.

The environment consists of a Unitree GO2 quadruped operating on sloped terrain with randomly parameterized rock obstacles. Experiments are conducted under varying obstacle orientations. We record the generated base goal positions, achieved base poses, end-effector trajectories, object motion trajectories, and contact force feedback measured at the end effector. We further compare the proposed method against two baselines, (1) a setup in which different goal poses are directly provided to the navigation policy, and (2) a setup in which alternative surface points are used as direct position goals for the pedipulation policy.

  \begin{table}[h]
\centering
\caption{Efficiency by Approach Angle Terrain Slope: \SI{225}{\degree}, Inward }
\label{tab:approach_efficiency}
\begin{tabular}{|l|l|l|l|}
\hline
\textbf{Heading (\SI{}{\degree})} & \textbf{Max Force (\SI{}{\newton})} & \textbf{Disp (\SI{}{\meter})} & \textbf{Efficiency (\SI{}{\meter\per\newton})} \\ \hline
(0-45)   &  63.46 & 0.0991 & 0.1561 \\
(45-90)  & 57.49 & 0.1035 & 0.1802 \\
(90-135) &  53.15 & 0.0968 & 0.1822 \\
(135-180)&  61.44 & 0.1998 & 0.3251 \\
(180-225)&  56.73 & 0.2035 & 0.3587 \\
(225-270)&  41.23 & 0.2020 & 0.4894 \\
(270-315)&  56.88 & 0.1557 & 0.2738 \\
(315-360)&  68.56 & 0.1442 & 0.2103 \\
\hline
\end{tabular}
\end{table}

For the first experiment type, the object is initialized with a random orientation, where the terrain slope vector forms an angle of \SI{225}{\degree} with respect to the world frame. The robot is randomly spawned around the object with an average distance of \SI{1}{\meter} with varying heading goals. For pedipulation, the closest point on the object surface to the robot base is selected in each trial to ensure consistency across evaluations. A total of 225 experiments are conducted, with heading targets varying in \SI{45}{\degree} increments to cover a full \SI{360}{\degree} range. Table~\ref{tab:approach_efficiency} shows the mean peak contact force applied to the object, the mean object displacement, and an efficiency metric defined as the ratio of displacement to peak force ($100*$\SI{}{\meter\per\newton}). The results demonstrate the effectiveness of the base pose affordance algorithm, which computes goal base positions by incorporating the local terrain slope around the object, thereby enabling more efficient object interaction through the end effector.

 \begin{figure}
    \centering
    \includegraphics[ width=0.5\textwidth, clip]{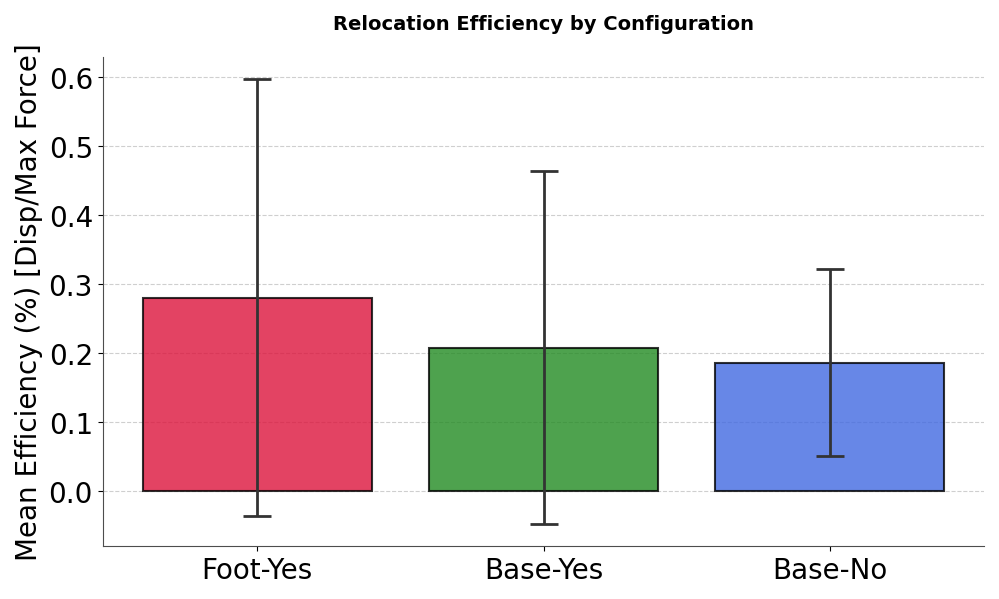}
    \caption{ Mean relocation efficiency across three interaction point selections (50 trials each). Efficiency is defined as displacement per maximum force (scaled by 100). The contact points selected as the closest points to the end-effector frame (Foot-Yes) achieve the highest efficiency, followed by the points selected based on their proximity to the robot base frame (Base-Yes) and the points chosen without any proximity-based selection criterion (Base-No). The error bars represent one standard deviation.}
    \label{fig:reach}
\end{figure}

The second experiment is conducted to analyze the effect of the selected pedipulation point on the object surface. Three different trajectory generation strategies are evaluated for the pedipulation policy: (1) selecting the closest surface point to the robot end effector, (2) selecting the closest surface point to the robot base, and (3) using the geometric center of the object without enforcing any specific surface contact point. A total of 150 experiments are conducted, with 50 trials per method. Figure~\ref{fig:reach} presents the efficiency scores corresponding to each choice of pedipulation point. The results show that selecting the closest point to the end effector achieves the highest efficiency, followed by the closest point to the base, while using no explicit surface contact point yields the lowest performance. These findings are consistent with the proposed pedipulation point selection strategy. The first row of Figure~\ref{fig:comp} illustrates a simulation experiment conducted on a flat surface, where the closest point to the end effector is used as the interaction point. The second row shows the corresponding real-world experiment performed with a different object.

\begin{figure*}
    \centering
    \includegraphics[ width=1.0\linewidth, trim={0 0 0 0}, clip]{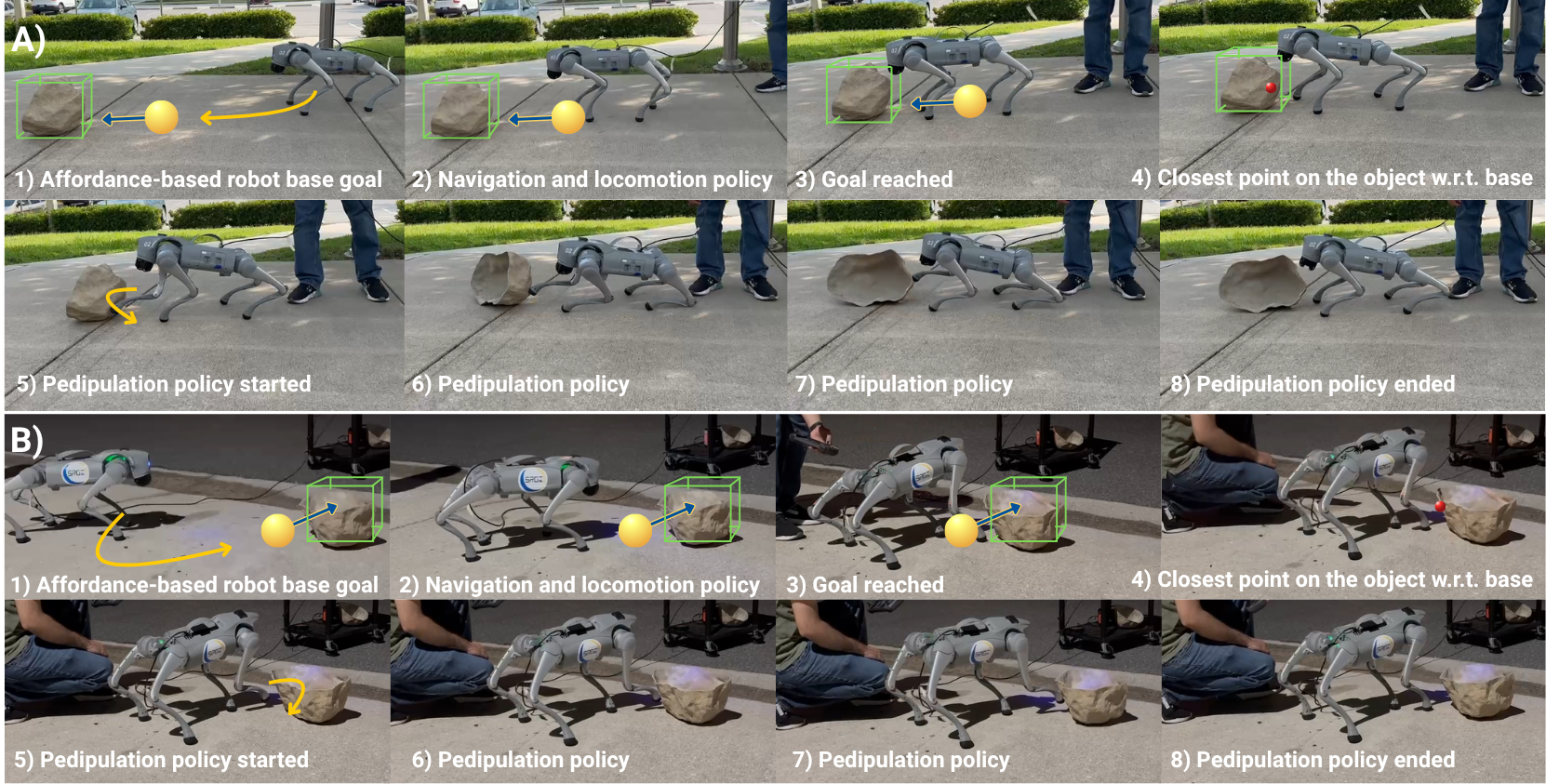}
    \caption{Real-world execution of the proposed system in open environments with varying surface slopes. First, the vision-guided pose affordance module generates a slope-aware goal pose for the robot base. The navigation module, conditioned on this goal pose, guides the locomotion policy toward the desired base position. After reaching the target within a predefined threshold, the system switches to the pedipulation phase, where the policy is conditioned on a geometry-aware interaction point on the object surface. The robot then performs the pedipulation task to manipulate the object.}
    \label{fig:realexp1}
\end{figure*}

\subsection{Real-World Experiments}
\label{real_world_section}

\begin{table*}[]
\centering
\caption{ Real World Experiments }
\label{table:real_world_experiments}

\begin{tabular}{|l|l|l|l|l|l|l|l|l|l|l|l|l|}
\hline
\# Exp & Slope &Pos Error &Yaw Error & Touch & WP0 Error & WP1 Error & WP2 Error & WP3 Error & Max Force & Object Disp & Efficiency  \\ \hline
Unit & (\SI{}{\degree}) &(\SI{}{\meter}) & (\SI{}{\degree}) & - & (\SI{}{\meter}) & (\SI{}{\meter}) & (\SI{}{\meter}) & (\SI{}{\meter}) & (\SI{}{\newton}) & (\SI{}{\meter}) & (\SI{}{\meter\per\newton}) \\ \hline
1  & 4.07 & 0.08 & 13.21 & YES & 0.03 & 0.07 & 0.11 & 0.02& 30 & 0.09 & 0.3   \\ 
2  & 4.23 & 0.18 & 2.96  & YES & 0.05 & 0.08 & 0.02 & 0.05 & 28 & 0.10 & 0.36   \\ 
3  & 4.64 & 0.10 & 12.94 & YES & 0.06 & 0.12 & 0.11 & 0.07 & 25 & 0.04 & 0.16   \\ 
4  & 5.07 & 0.02 & 8.62  & YES & 0.12 & 0.15 & 0.03 & 0.12 & 29 & 0.04 & 0.14   \\ 
5  & 5.12 & 0.15 & 23.01 & YES & 0.06 & 0.12 & 0.03 & 0.06 & 48 & 0.36 & 0.75   \\ 
6  & 6.11 & 0.14 & 22.86 & YES & 0.01 & 0.09 & 0.04 & 0.01 & 36 & 0.05 & 0.14   \\ 
7  & 6.46 & 0.22 & 10.21 & YES & 0.04 & 0.15 & 0.07 & 0.04 & 42 & 0.58 & 1.38   \\ 
8  & 6.76 & 0.02 & 0.49  & YES & 0.14 & 0.21 & 0.08 & 0.14 & 55 & 0.04 & 0.07   \\
9  & 7.30 & 0.15 & 1.72  & YES & 0.01 & 0.11 & 0.03 & 0.01 & 35 & 0.18 & 0.51   \\ 
10 & 8.06 & 0.09 & 22.42 & YES & 0.02 & 0.07 & 0.05 & 0.02 & 28 & 0.69 & 2.46   \\
11 & 8.06 & 0.09 & 22.90 & NO  & 0.00 & 0.05 & 0.15 & 0.00 & 24 & -    & -      \\
12 & 8.09 & 0.11 & 31.74 & YES & 0.01 & 0.10 & 0.01 & 0.01 & 26 & 0.05 & 0.19   \\ 
13 & 8.19 & 0.29 & 37.71 & YES & 0.00 & 0.10 & 0.05 & 0.00 & 31 & 0.12 & 0.39   \\ 
14 & 9.14 & 0.00 & 18.42 & NO  & 0.03 & 0.09 & 0.07 & 0.03 & 22 & -    & -      \\ \hline
\end{tabular}
\end{table*}

\subsubsection{State Estimation}
The observation space of our policies requires the base position, linear velocity, and height expressed in the world frame. To estimate these quantities, we employ an Error State Extended Kalman Filter (ESEKF) to fuse high-frequency IMU predictions with kinematic measurements. During locomotion, each leg alternates between swing and stance phases, independent of the specific gait. The fundamental assumption of the leg odometry framework is that the foot velocity of leg $i$ relative to the world frame is zero during the stance phase, i.e., $^\mathcal{W}\bm{v}_{\text{foot}, i} = \bm{0}$. Consequently, leg odometry frameworks treat the stance phase as a pseudo-measurement to correct the estimated base linear velocity via the leg's kinematic chain.

While the state variables are propagated forward in time using IMU integration, a measurement update is triggered whenever a leg is detected in the contact phase. Therefore, the measurement model is defined as follows:

\begin{equation}
    h(\bm{x}_k) = ^\mathcal{W}\!\!\bm{v}_k + \bm{R}_k {^\mathcal{B}\!\bm{v}_{\text{rel},i}},
\end{equation}

\noindent where $^\mathcal{W}\bm{v}_k$ is the base linear velocity in the world frame, $\bm{R}_k$ is the rotation matrix from the body frame to the world frame, and $^\mathcal{B}\bm{v}_{\text{rel},i}$ is the foot velocity relative to the base, derived via the kinematic transport theorem at time $k$. Given the pseudo-measurement $\bm{y}_k = \bm{0}$, the innovation used to compute the Kalman gain and update the state vector is $\bm{\nu}_k = -h(\bm{x}_k)$.

To detect stance phases, we adopt the OCELOT framework \cite{girgin2026ocelotodometrycontactestimation}, which combines physical load modeling represented as a bimodal Gaussian based on ground reaction forces with a Generalized Likelihood Ratio Test (GLRT) applied to the estimated foot velocity. Refer to \cite{girgin2026ocelotodometrycontactestimation} for further details.

\subsubsection{Vision}
To dynamically detect the target object, generate the navigation goal, and determine the initial surface contact point, we implement a point cloud-based clustering and normal estimation framework adapted from \cite{girgin2025learning}. Point clouds are acquired using the onboard, body-mounted rotating LiDAR and transformed into the base frame. A $k$-d tree algorithm processes the point cloud to estimate local surface normals. Assuming the median vector of all computed normals represents the reference ground normal, we calculate the cosine similarity between each local normal and this reference. Points yielding a cosine similarity above a specified threshold (0.88) are classified as ground, and the remainder are assumed to belong to an object resting on the surface. We then apply the DBSCAN algorithm to the non-ground points to extract discrete object clusters. Although this framework supports multi-object detection, we restrict our experiments to a single object in the field of view to bypass high-level target selection logic.

After computing the 3D bounding box of the object cluster, the local terrain slope is estimated by averaging the normal vectors of the ground points located within the bounding box's footprint. The navigation goal is defined along the surface gradient (direction of steepest ascent), applying a robot morphology-dependent translational offset from the object's geometric center keeping the object in robot's reach. Finally, the contact start position ($\bm{p}_{push}$) is defined as the object surface point closest to the robot's base frame. This selection maximizes the available contact trajectory length, thereby enabling maximum displacement of the object.

\subsubsection{Sim-to-Real}
The proposed framework was deployed on the Go2 quadruped using the open-source ROS2 deployment framework \texttt{\href{https://github.com/eppl-erau-db/go2\_rl\_ws}{go2\_rl\_ws}}. A finite state machine (FSM) manages transitions across eight operating modes: Idle, Standing, Sitting, Walking, Pedipulation, EmergencySitting, Damping, and Killed, selecting the active mode based on the robot's current posture and the requested transition command. During hardware testing, wireless controller buttons were used solely to request or confirm mode changes, such as arming motion, entering pedipulation, triggering the push motion, or requesting a safe stop, rather than to steer the robot or manually define the task trajectory. These inputs served only as safety checks and transition confirmations, and can be replaced by autonomous software triggers in fully autonomous operation. A posture monitor evaluates joint pose status to inform FSM transitions and enforce hard joint limit constraints, overriding the FSM in emergency conditions to prevent hardware damage. Policy inference is performed using ONNX models and ONNX Runtime~\cite{onnxruntime}, enabling cross-platform deployment. Because inference was offloaded to an external computer connected via Ethernet, the resulting latency is negligible relative to the task timescale.


\subsubsection{Testing}

For the real-world experiments, we conducted tests on concrete terrain with varying slopes, where faux rocks were randomly repositioned and reoriented, with an average distance of 1 m from the robot. As shown in Table~\ref{table:real_world_experiments}, a total of 14 real-world trials were performed across surfaces with different estimated slope conditions. Across all real-world experiments, the Go2's built-in locomotion policy was used due to its increased robustness characteristics. Despite being trained on our simulated locomotion policy, the navigation policy generalizes well to the domain shift. Guided by the navigation policy, the robot base reached the desired goal pose with a mean position error of \SI{0.12}{\meter} and a mean yaw error of \SI{16.37}{\degree}.

Since the pedipulation trajectory is defined as a cubic spline over four waypoints, we evaluate performance by measuring the minimum distance between the end effector and each intermediate waypoint during a full execution cycle. After the locomotion phase, the system transitions directly to the pedipulation policy, resulting in a randomized initial joint state at the start of manipulation. From these initial conditions, the end effector achieves a mean error of \SI{0.04}{\meter} at waypoint 0,  \SI{0.10}{\meter} at waypoint 1,\SI{0.06}{\meter} at waypoint 2, and \SI{0.04 }{\meter} at waypoint 3. The waypoints correspond to equally spaced points along the trajectory, including the start and end points. The system achieves an object interaction success rate of 86\% from the reached base poses, where the interaction threshold force is defined as \SI{25}{\newton}, extracted from the force readings when the end effector is idle. Across trials, the system achieves a mean object displacement of \SI{0.17 }{\meter}, a mean peak contact force of  \SI{33 }{\newton}, and a mean efficiency score of \SI{0.51 }{\meter\per\newton}.

Figure~\ref{fig:realexp1} illustrates real-world scenarios in which the GO2 robot reaches the navigation goal and interacts with a randomly positioned object using closest-point, offset-point, and home-position-based pedipulation trajectories, respectively.

\section{CONCLUSIONS}
\label{sec:conclusion}

We proposed a three-level hierarchical reinforcement learning framework  that autonomously detects target objects and generates affordance-aware target base poses for initiating pedipulation. Using a navigation policy–guided locomotion policy, the GO2 robot reaches the commanded base position. Subsequently, the system identifies an affordable interaction point on the object surface to maximize the efficiency of object relocation. The pedipulation policy is then employed to generate end-effector commands for physical interaction with the object, maintaining the balance with the remaining three legs. The proposed framework is evaluated both in simulation and in real-world environments, where interaction data is collected during physical experiments. This work enables exploration scenarios such as sample collection and site inspection in previously unseen environments, requiring the identification of suitable operating poses for the robot base and end effector under varying environmental conditions, while enabling dexterous manipulation using highly maneuverable legged robots without additional hardware.
A key challenge in this setting is that the system comprises multiple interconnected components, including vision perception, state estimation, navigation, locomotion, and pedipulation policies. Errors accumulated across these stages can significantly affect overall task success, particularly due to the high precision required for dexterous object manipulation.

For future work, we aim to leverage the collected interaction data to learn object-centric representations that can be used to improve downstream manipulation and in-situ utilization of objects.

\addtolength{\textheight}{-0cm}   




\bibliographystyle{IEEEtran}
\bibliography{references}

\end{document}